\documentclass{article}
\usepackage{spconf,amsmath,graphicx}
\usepackage{times}
\usepackage{url}
\usepackage{latexsym}
\usepackage{paralist}
\usepackage{multirow}
\usepackage{amssymb,verbatim}

\title{Modeling Melodic Feature Dependency with \\ Modularized Variational Auto-Encoder}
\name{Yu-An Wang$^\star$, Yu-Kai Huang$^\star$, Tzu-Chuan Lin$^\star$, Shang-Yu Su, Yun-Nung Chen}
\address{National Taiwan University, Taipei, Taiwan\\
\texttt{\small \{b04902004, b04902131, b04705003, f05921117\}@csie.ntu.edu.tw\quad y.v.chen@ieee.org}}
\begin{document}
\ninept
\maketitle
\begin{abstract}
Automatic melody generation has been a long-time aspiration for both AI researchers and musicians.
However, learning to generate euphonious melodies has turned out to be highly challenging.
This paper introduces 1) a new variant of variational autoencoder (VAE), where the model structure is designed in a modularized manner in order to model polyphonic and dynamic music with domain knowledge, and 2) a hierarchical encoding/decoding strategy, which explicitly models the dependency between melodic features. The proposed framework is capable of generating distinct melodies that sounds natural, and the experiments for evaluating generated music clips show that the proposed model outperforms the baselines in human evaluation. \footnote{The first three authors contributed equally.}
\end{abstract}

\section{Introduction}
Recently, in algorithmic music generation field, there are two mainstreams: \textit{symbolic-domain} and \textit{audio-domain}. In symbolic-domain, its target is to generate music in standard MIDI format\cite{perf_rnn, multi-musicvae, yang2017midinet, dong2018convolutional, jambot, deepj, biaxial}. On the other hand, in audio-domain, its goal is to synthesize music that sounds realistic\cite{Wavenet, nsynth_wavenet, samplernn}. In this paper, we focus on \textit{symbolic-domain} music generation.

Generating music is different from generating other modalities like images or natural languages. To create harmonic music, the order and combination of music elements in the temporal scale are of significant importance.
There are two main modeling directions for music generation, one uses sequence modeling such as recurrent neural networks (RNN)\cite{perf_rnn,  jambot, deepj, model_temporal, biaxial}, and another uses generative modeling such as variational autoencoders (VAE) ~\cite{VAE_paper} and generative adversarial networks (GAN)~\cite{MuseGAN, yang2017midinet}.
To model the sequence-like attributes of music elements, we utilize variation auto-encoders (VAE) to generate polyphonic and dynamic music.
Further extension of MusicVAE - Multitrack MusicVAE~\cite{musicvae} and SeqGAN~\cite{seqgan} was proposed to focus on multi-track music.


\begin{figure*}[t!]
	\centering
    \centerline{\includegraphics[width=0.95\linewidth]{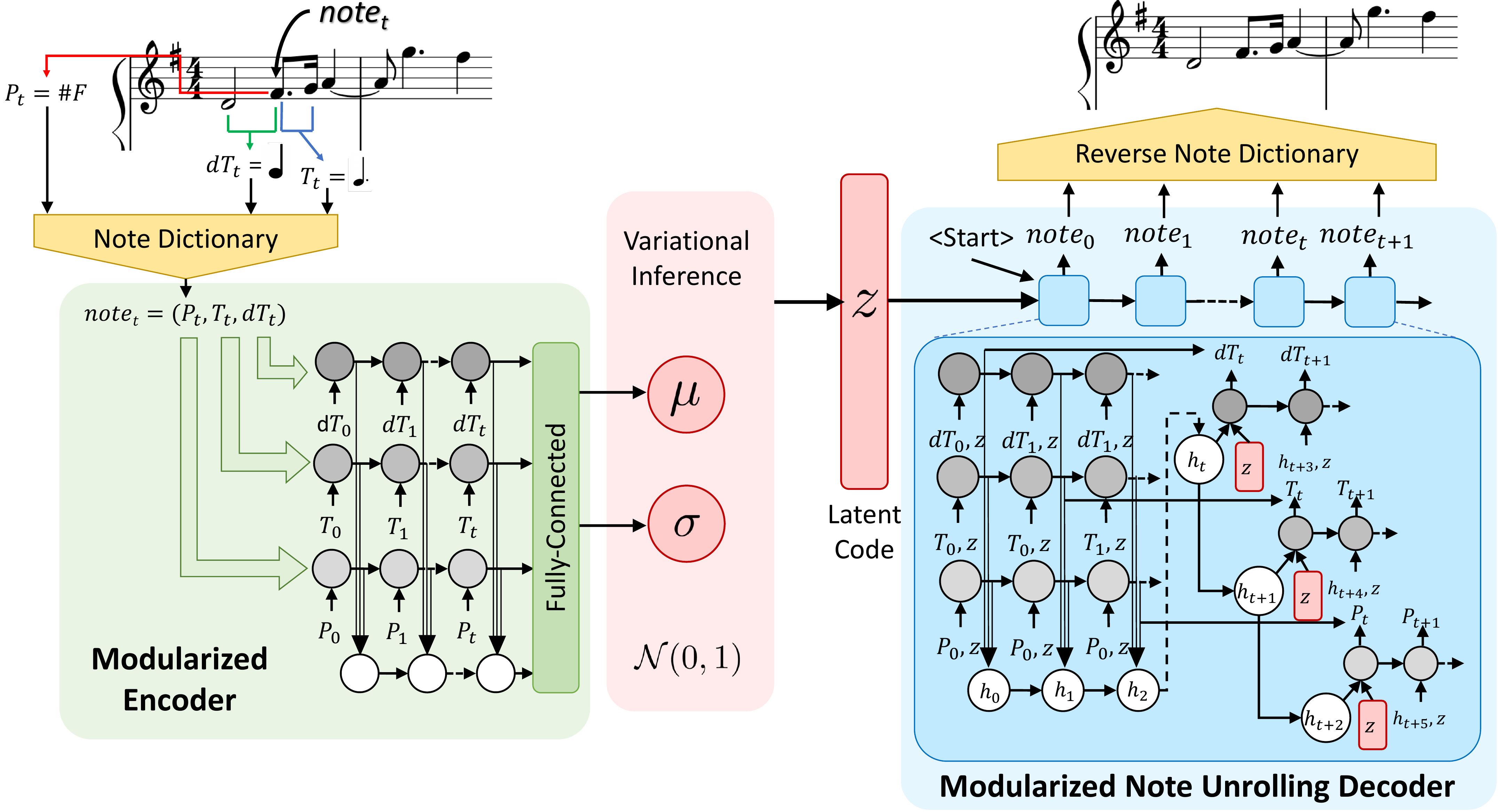}}
    \vspace{-2mm}
	\caption{The illustration of the proposed modularized framework, where the VAE architecture is embedded with a modularized encoder and a note-unrolling decoder, and the note event representations are realized by a note dictionary and a reverse note dictionary.}
	\label{fig:model}
    \vspace{-3mm}
\end{figure*}

Because music is often seen as sequential information, RNN is applied to music generation.
In this work, we utilized an integrated model, variational recurrent auto-encoders (VRAE)~\cite{VRAE}, incorporating the benefits of modeling long-term dependencies by recurrent units and generative nature of VAE, and further introduce a novel generative model for music generation.
First, we propose the architecture of encoder so as to encode more information into the latent code.
With domain knowledge, we modularize the encoder into two parts: the first part mainly focuses on the rhythm and the pitch of a note, while the second part gathers information and puts the essence of the first part into contexts.
Moreover, we use a hierarchical and recurrent approach, note unrolling, to model the dependency of music notes.
This is the first work applying note unrolling in VAE model, and we find that it is suitable for the decoder to explicitly model the dependency of time, duration and pitch of music.
Our main contributions are as follows:
\begin{compactitem}
	\item The proposed model incorporates domain knowledge of music by using a modularized framework for modeling various melodic features.
    \item This is the first work that well integrates the note-unrolling technique in VAE to model the dependency between melodic features for music generation.
    \item The proposed model is capable of generating natural music from the human perspective and achieves better performance than other generative models.
\end{compactitem}


\section{Proposed Model}

In the proposed model, the gated recurrent unit (GRU)~\cite{GRU} is applied to formulate variation recurrent auto-encoders (VRAE)~\cite{VRAE}, considering its balance between performance and model size~\cite{empirical_GRU}.
For each datapoint $x$, the training object of VAE is evidence lower bound objective (ELBO),
\begin{align}
\mathcal{L}_\text{ELBO}(x) &= \mathbb{E}_{q_{\phi}(z \mid x)}[ \log p_{\theta}(x\mid z)] - \mathit{D}_{KL}(q_{\phi}(z \mid x) \parallel p_{\theta}(z)).\nonumber
\end{align}
The first term of ELBO can be viewed as the reconstruction loss; the second one is the regularization term, which is the Kullback-Leibler (KL) divergence between the amortized inference distribution and the prior $p_{\theta}(z)$.
The choice of these distribution is often a factorized Gaussian by its simplicity and computational efficiency. 
This work utilizes the normal distribution with a diagonal covariance matrix, $p_{\theta}(z) = \mathcal{N}(0, I)$. 
The whole objective can be optimized by gradient-based methods and reparametrization tricks with respect to the parameters $\phi$ and $\theta$.

The encoder, $q$ first encodes an input sequence $ \mathbf{x} = (x_1, \cdots, x_T)$ as a normal distribution $\mathcal{N}(z \mid \mu(x, \phi), \Sigma(x, \phi))$.
Then the RNN decoder generates an output sequence given the sampled latent vector $z$ from the normal distribution of the encoder.
Different from conventional sequence modeling as previous described, VRAE can effectively learn to represent an output sequence due to the objective constraints on the KL divergence.
Therefore, VRAE is able to embed richer semantic information into the latent space $Z$ than traditional sequence modeling approaches.
This paper builds a music-generation framework on top of VRAE and allows us to incorporate domain knowledge for the target task by the modular components.

The proposed framework integrates the advantages from the prior work and proposes a novel model with better flexibility and performance, which is illustrated in Figure~\ref{fig:model}. 
Considering that BachProp~\cite{bachprop} used the normalized note representation of music, we uses similar note representation, called note event representation, which is capable of translating music with minimal distortion. 

In the previous work, MusicVAE~\cite{musicvae} utilized VAE models to generate music. Moreover, MusicVAE applied a hierarchical decoder on each measure so that MusicVAE can learn the long-term structure of music. 
However, the hierarchical framework was limited to generate only 4/4 time music piece due to its decoder's structure.
In contrast, our model is not constrained to generate specific time signature music by maintaining the structure between notes by note unrolling, demonstrating better flexibility and practical usage.

\subsection{Data Representation}

In order to represent more dynamic and complex music, this paper proposes to use the \emph{note event} as a basic unit of music, which can separate into three attributes: 
\begin{equation}
\{\mathbf{note}_t=(dT_t,T_t,P_t) \},
\end{equation}
where $t$ indicates $t$-th note event, $dT$ represents starting time difference of note events between $\mathbf{note}_{t-1}$ and $\mathbf{note}_t$, and $T$, $P$ represent the duration and pitch of one note event, respectively. 
By setting the value of $dT_t=0$, the note $\mathbf{note}_t$ and the previous note $\mathbf{note}_{t-1}$ can be pressed at the same time, producing a polyphonic structure like chords and mixed chorus. 
The top-left part of Figure~\ref{fig:model} illustrates the note dictionary.

Previous work, such as MusicVAE~\cite{musicvae} and DeepBach~\cite{DeepBach}, discretized time into sixteenth notes. 
However, note event representation allows the model to learn beats which are not multiples of sixteenth notes and have the freedom of pressing multiple notes at same time to generate polyphonic music. 
The attributes are indexed by the appearance order in dataset.

\subsection{Shared Embedding}

From the human knowledge of music theory, we know different beats and pitches have certain relationships. For example, C, G and B, \#F are in C and B major, respectively.
A triplet quarter note with duration $\frac{1}{6}$ and three sixteenth notes with duration $\frac{3}{16}$ have close duration value.
In this work, we project the discrete representation of each melodic feature into a embedding space to model these relationships before feeding into the encoder and the decoder.
Empirically, using shared embeddings for each feature in the encoder and the decoder can reduce parameters while having same performance. 

\subsection{Modularized Encoder}
Na\"ively, we can use only one large GRU taking the concatenation of $dT_t$, $T_t$ and $P_t$'s embedding vector as input.
However, rhythm and pitch are different category of attributes in music theory, which means it is possible that once encoder gets rhythm mixed up with pitch-related information, it could not understand this music piece well.
As a result, we propose a modularized encoder encouraging them to focus on encoding different melodic features separately and extracting cleaner information.

The proposed modularized encoder is illustrated in Figure \ref{fig:model}, which consisting of four GRUs. 
The first three GRU encoders extract respective messages from different melodic features: time($dT$), duration($T$) and pitch($P$), without inter-connections among one another. 
To combine the extracted information from these encoders, the context GRU takes output vectors from the above three encoders at each time step as input and integrates information.
Further, we concatenate the the latent vectors at the last step from all four GRU encoder modules and pass it through fully-connected layers to get the variational parameters of the distribution over latents, which is the mean $\mu$ and standard deviation $\sigma$. 

Formally, the proposed modularized encoder can be represented as following mathematical forms:
\begin{align}
	h_{t}^{dT} &= \mathbf{GRU}_{dT}(h_{t-1}^{dT}, dT_t), \nonumber \\
    h_{t}^{T} &= \mathbf{GRU}_{T}(h_{t-1}^{dT}, T_t), \nonumber \\
    h_{t}^{P} &= \mathbf{GRU}_{P}(h_{t-1}^{P}, P_t), \nonumber \\
    h_{t}^{C} &= \mathbf{GRU}_{C}(h^{C}_{t-1}, [h_{t}^{dT}; h_{t}^{T}; h_{t}^{P})), \nonumber
\end{align}
where $\mathbf{GRU}_C$ is the context GRU that gathers information from the lower-level modules  $\mathbf{GRU}_{dT}$, $\mathbf{GRU}_{T}$ and $\mathbf{GRU}_{P}$.
Finally, we have $\mu$ and $\sigma$ by transformation with a few linear layers:
\begin{eqnarray}
	v &=& W_{1}[h_{T}^{dT}; h_{T}^{T}; h_{T}^{P}; h_{T}^{C}] +  b_{1}, \\
    \mu &=& W_{\mu}v + b_{\mu}, \nonumber \\
    \sigma &=& \exp(\frac{1}{2}(W_{\sigma}v + b_{\sigma})). \nonumber
\end{eqnarray}

\subsection{Modularized Note-Unrolling Decoder}
In the field of music theory, rhythm, modes and tone are combined with dependency to create a melody. Rhythm stands for time and duration to some degree, while the lowness and highness of a tone represents pitch.
All attributes of a note for a chord ($[dT, T, P]$) are combined with some relation to fit that chord.
Considering the nature of music composition described above, we should model the dependency between attributes of chords explicitly while decoding. 

The previous work \cite{bachprop} models the dependency relations between $dT$, $T$ and $P$ by decomposing the joint probability of three attributes in a note event into a product of conditional probabilities, which can be written as:
\begin{eqnarray}
  p(dT_t, T_t, P_t) &=& p(dT_t\mid \mathbf{note}_{1:t-1}) \\
  &\times& p(T_t \mid dT_t, \mathbf{note}_{1:t-1})  \nonumber\\
  &\times& p(P_t \mid dT_t, T_t, \mathbf{note}_{1:t-1}). \nonumber
\end{eqnarray}
The design of note unrolling follows music domain knowledge, where the note attributes are often conditioned on other attributes.
Figure \ref{fig:model} illustrates the concept of note unrolling and the design of the proposed hierarchical decoder, which contains total 7 GRUs: three for modeling attribute-specific contexts, one for combining multiple attributes as a contextual module, and three for generating associated note attributes.
Furthermore, we utilize residual skip connection from upper-level modules to lower-level modules, since our modularized decoder have many GRUs, skip connection can avoid gradient vanishing caused by back propagation through too many layers.
The hierarchical decoder separates the generation procedure into subsequent three process: time, duration and pitch; 
it is taught to output $dT_{t}$ depending on previous notes $\mathbf{note}_{1:t-1}$.
After that, the second step depends on $dT_{t}$ and $\mathbf{note}_{1:t-1}$ to generate $T_{t}$.
Finally, the network depends on $dT_{t}$, $T_{t}$ and $\mathbf{note}_{1:t-1}$ to generate $P_{t}$.

\subsection{Training and Generation}
We optimize our model by RMSprop optimizer with learning rate $10^{-4}$ and a batch size of 128.
Due to strong autoregressive characteristics in RNN, VRAE tends to ignore the latent distribution, so-called \emph{posterior collapse} issue.
To mitigate this problem, KL annealing is applied~\cite{Sentence_VAE} to allow the model to encode more information into the latent code $\mathbf{z}$ at first and then gradually fit the prior as the weight approached $1$.

In the inference stage, we sample $k$-dimension latent code $\mathbf{z}$ from the standard normal distribution $\mathcal N (0,I_k)$ as input of the decoder, where $k$ indicates the dimension of latent vector. 
Next, we generate music pieces of length 100 notes by the proposed hierarchical decoder with all the techniques mentioned above.  

\section{Experiments}

\subsection{Setup}
The experiments are performed on three diverse benchmark datasets: Nottingham, Piano-midi.de and JSB Chorales, which are frequently used for music generation~\cite{boulanger2012modeling,bachprop}.
Nottingham dataset contains 1037 midi files of folk music; 
Piano-midi.de dataset contains 333 midi files of classical music; 
JSB Chorales dataset contains 382 midi files of chorales of J.S Bach.
To better validate the capability of modeling the music diversity in our generative model, we merge three into one large dataset for training. The hidden size of all GRUs in our model is set to 512.
Considering the difficulty of modeling long songs in the dataset, the midi files are cut into 100 note segments with stide equal to 50. Then we randomly rearrange tonality of each segment by $[-3, +3]$ for data augmentation.



\subsection{Baseline}
We compare our proposed model two baselines, BachProp~\cite{bachprop} and modularized autoencoder.

\begin{compactitem}
\item BachProp: the model similar to our proposed model without an encoder and variational approximation on latent distribution, hence the comparison can highlight the importance of our modularized encoder. 
\item Modularized autoencoder: the model similar to our model without the objective of KL divergence constraint on the latent distribution $z$, which could show the importance of the variational inference.
\end{compactitem}
Note that we do not compare with MusicVAE~\cite{musicvae}, because it cannot handle the music that is not 4/4 time signature contained by our dataset, making the comparison inapplicable.
Another advantage of our model is the flexibility of modeling dynamic music with different time signatures compared with MusicVAE.

\subsection{Human Evaluation}
To measure the performance of the proposed model, we conduct human evaluation, the procedure is designed as following.
First, the recruited raters are asked about their music background: small, medium, or strong. 
Second, the raters are requested to give the Likert scale scores from 1 to 6 to measure whether the music is human-composed (higher score) or machine-generated (lower score) for each given 100-note midi file.
Finally, We collect 85 scores for each model and dataset.
\begin{table}[t!]
\centering
\caption{Experimental results of reconstruction loss (\textbf{Rec.}), KL divergence loss (\textbf{KL}) for each model with different settings. }
\label{tab:training_results}
\vspace{2mm}
\begin{tabular}{ | l  r r | c  c|}
    \hline
    \multirow{2}{*}{\bf Model} & \multirow{2}{*}{\bf Rec.} & \multirow{2}{*}{\bf KL}& \multicolumn{2}{|c|}{\bf Human Score}\\
    & & & \(\mu\)& \(\sigma\)\\
\hline \hline
BachProp~\cite{bachprop}          & 240.16 & --- & 3.51 & 1.61\\
Modularized autoencoder        & 20.79 & --- & 2.77& 1.65\\
\hline
\multicolumn{5}{|l|}{\emph{Proposed model}} \\
~~~w/o note unrolling & 85.88 & 264.00 & 3.22$^\dag$ & 1.73\\
~~~w/ note unrolling & 73.19 & 30.37 & 4.24$^{\dag\ddag}$ & 1.54\\
\hline
Real Data & --- & --- & 4.34 & 1.55\\
    \hline
  \end{tabular}
\end{table}

\subsection{Results}
The human evaluation results are shown in Table~\ref{tab:training_results}, where we perform the significance test to validate the improvement.
The improvement achieved by our model compared to BachProp and modularized autoencoder is both statistically significant using the single-tailed t-test with $\alpha < 0.01$ marked as $^\dag$.

It tells that our model is capable of generating music according to the designated encoded $\mathbf{z}$ and, thus, has more flexibility and diversity. 
In contrast with BachProp, it has no information from the latent code $\mathbf{z}$ and cannot maintain a consistent structure in an output sequence. Thus, its scores are much lower than our proposed model by a large margin.  
When comparing with modularized autoencoders, it hard codes $\mathbf{z}$ for each song in the dataset on the latent space.
However, without KL loss during training, the decoder cannot model the meaning of a randomly sampled $\mathbf{z}$ at the inference phase.
The human evaluation results show that our proposed model can obtain an informative latent code $\mathbf{z}$ using VAE and outperform other baseline generative models.
The achieved performance is close to the real data scores, which are the upperbound of our music generation model.

\subsection{Effectiveness of Note-Unrolling Decoder}
The note-enrolling mechanism was first proposed by BachProp on the sequence prediction model, and they claimed that there is dependency between $dT, T$ and $P$ for one note according to human knowledge~\cite{bachprop}. 
However, they did not illustrate or analyze how note unrolling is better than original approaches. Therefore, we perform an ablation test to explicitly verify whether the note unrolling decoder brings the benefit on generating pleasant music pieces. 
From Table~\ref{tab:training_results}, adding note unrolling can decrease the reconstruction loss and significantly improve the human scores.
The difference between their results is significant with $\alpha<0.01$ marked as $^\ddag$, demonstrating the effectiveness of modeling music structures based on music domain knowledge in our proposed model.


\subsection{Analysis of Latent Space}
Our dataset is composed of multiple music characteristics including folk music, classical music, and chorales of J.S Bach.
In order to check 1) whether our model can learn diverse characteristics in the same latent space and 2) whether our model can capture the difference about characteristics and encode into the informative codes, we perform the following analysis.

\begin{figure}[t!]
    \centering
    \includegraphics[width=0.9\columnwidth]{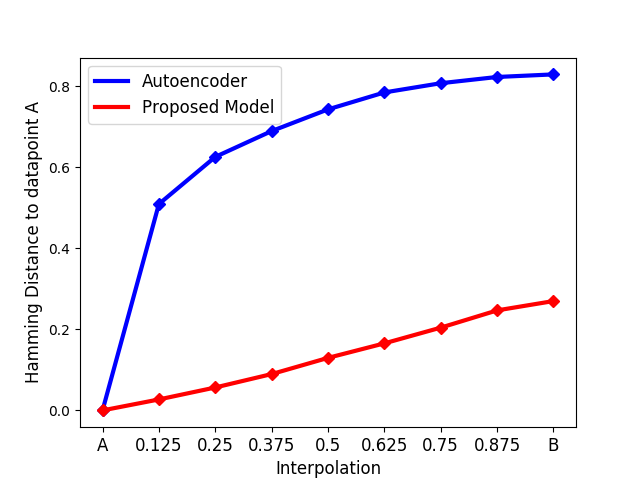}
    \vspace{-4mm}
    \caption{Average distance between two random datapoints on $Z$.}
    \label{fig:hd_distance}
    \vspace{-4mm}
\end{figure}

\subsubsection{Interpolation Distribution}
To further analyze whether our proposed approach is capable of modeling diverse music characteristics, we sample two datapoints, A and B, compute the interpolation points between them using their latent codes, and further check whether our model can smoothly model their distribution.
We varify the smoothness of the latent space distribution by computing Hamming distance between every interpolation points and datapoints.
Undoubtedly, we find that as the interpolation point goes from data point A to data point B, its Hamming distance to A starts increasing but decreasing in distance to B. Note that we only show the Hamming distance to A due to the tendency of two curves is almost symmetric.

The results are shown in Figure \ref{fig:hd_distance}, where our proposed model has a more smooth curve than the baseline autoencoder, implying that the interpolation points between two data points are meaningful.
To sum up, our proposed model can handle the diverse inputs by constraining the distribution using KL, while the autoencoder models diverse characteristics in multiple separate spaces.


\subsubsection{Visualization}

\begin{figure}[t!]
  \centering
  \includegraphics[width=0.95\columnwidth]{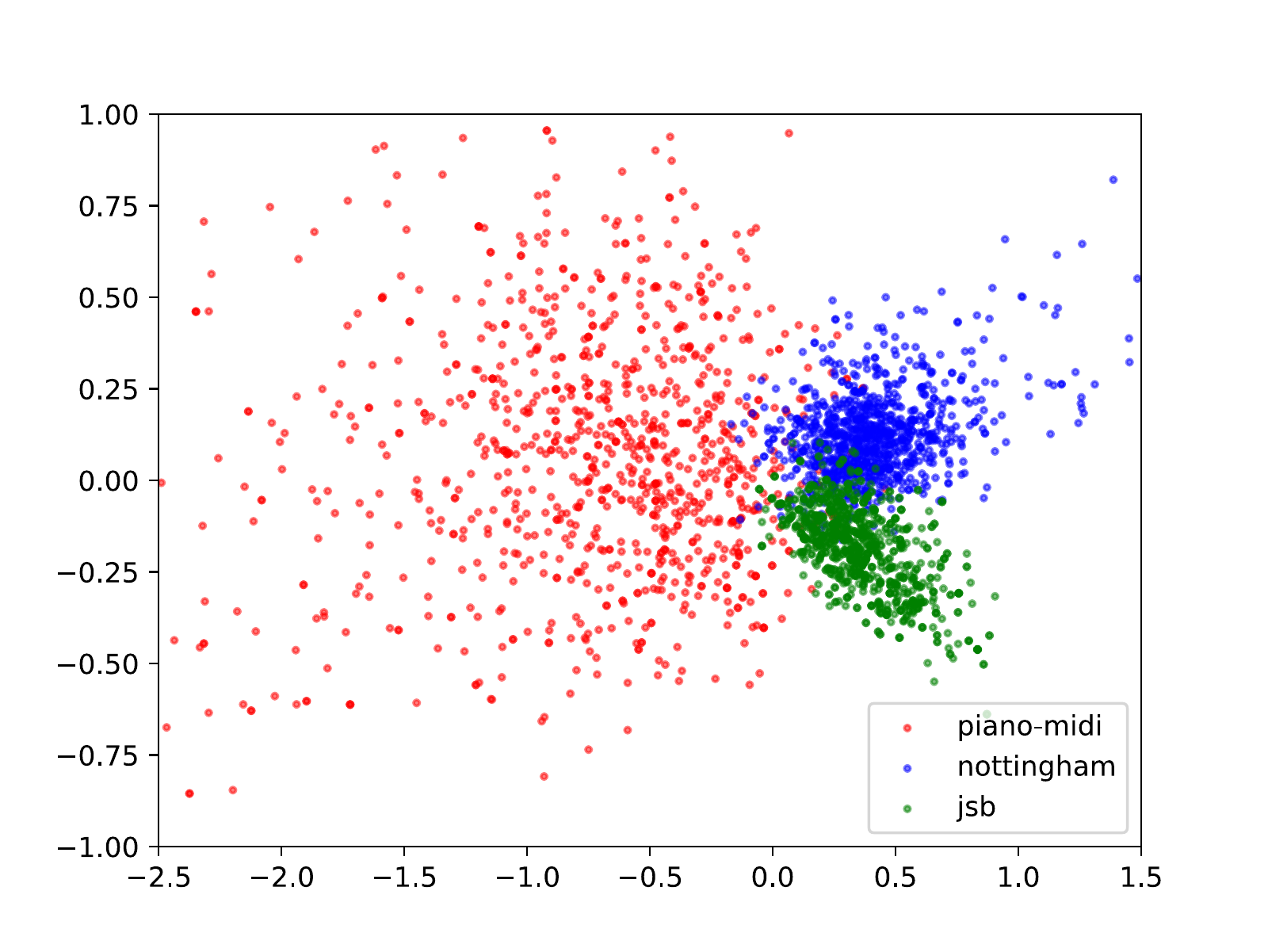}
  \vspace{-5mm}
  \caption{Visualization on the latent space via PCA, where three different types of music are separated in $Z$.}
  \label{fig:z_visual}
  \vspace{-4mm}
\end{figure}

To further analyze whether our mode can effectively capture the distinct features among different characteristics in music, we project the learned latent codes of different datasets into 2-dimensions by PCA.
The results are shown in Figure \ref{fig:z_visual}, where $\mathbf{z}$ of different type of music are separated in the latent space.
The result demonstrates that our encoder can surely encode music into a informative latent code.

\section{Conclusion}
This paper presents a VAE model that incorporates a modularized encoder and a modularized decoder in the framework to better generate realistic melodies.
The modularized encoder is capable of encoding the latent information, and a note unrolling decoder models the melodic dependency between note attributes. 
Also, the proposed note event representations bring the better flexibility.
The experiments are conducted in a merged dataset with diverse characteristics in music, demonstrating the superior performance of our proposed model for all evaluation scenarios: human evaluation and latent space analysis.

\vfill\pagebreak

\bibliographystyle{IEEEbib}
\bibliography{strings,refs}

\begin{thebibliography}{10}

\bibitem{perf_rnn}
Ian Simon and Sageev Oore,
\newblock ``Performance rnn: Generating music with expressive timing and
  dynamics,'' 2017.

\bibitem{multi-musicvae}
Ian Simon, Adam Roberts, Colin Raffel, Jesse Engel, Curtis Hawthorne, and
  Douglas Eck,
\newblock ``Learning a latent space of multitrack measures,''
\newblock {\em arXiv preprint arXiv:1806.00195}, 2018.

\bibitem{yang2017midinet}
Li-Chia Yang, Szu-Yu Chou, and Yi-Hsuan Yang,
\newblock ``Midinet: A convolutional generative adversarial network for
  symbolic-domain music generation,''
\newblock {\em arXiv preprint arXiv:1703.10847}, 2017.

\bibitem{dong2018convolutional}
Hao-Wen Dong and Yi-Hsuan Yang,
\newblock ``Convolutional generative adversarial networks with binary neurons
  for polyphonic music generation,''
\newblock {\em arXiv preprint arXiv:1804.09399}, 2018.

\bibitem{jambot}
Gino Brunner, Yuyi Wang, Roger Wattenhofer, and Jonas Wiesendanger,
\newblock ``Jambot: Music theory aware chord based generation of polyphonic
  music with lstms,''
\newblock in {\em Tools with Artificial Intelligence (ICTAI), 2017 IEEE 29th
  International Conference on}. IEEE, 2017, pp. 519--526.

\bibitem{deepj}
Huanru~Henry Mao, Taylor Shin, and Garrison Cottrell,
\newblock ``Deepj: Style-specific music generation,''
\newblock in {\em Semantic Computing (ICSC), 2018 IEEE 12th International
  Conference on}. IEEE, 2018, pp. 377--382.

\bibitem{biaxial}
Daniel~D Johnson,
\newblock ``Generating polyphonic music using tied parallel networks,''
\newblock in {\em International Conference on Evolutionary and Biologically
  Inspired Music and Art}. Springer, 2017, pp. 128--143.

\bibitem{Wavenet}
Aaron van~den Oord, Sander Dieleman, Heiga Zen, Karen Simonyan, Oriol Vinyals,
  Alex Graves, Nal Kalchbrenner, Andrew Senior, and Koray Kavukcuoglu,
\newblock ``Wavenet: A generative model for raw audio,''
\newblock {\em arXiv preprint arXiv:1609.03499}, 2016.

\bibitem{nsynth_wavenet}
Jesse Engel, Cinjon Resnick, Adam Roberts, Sander Dieleman, Douglas Eck, Karen
  Simonyan, and Mohammad Norouzi,
\newblock ``Neural audio synthesis of musical notes with wavenet
  autoencoders,''
\newblock {\em arXiv preprint arXiv:1704.01279}, 2017.

\bibitem{samplernn}
Soroush Mehri, Kundan Kumar, Ishaan Gulrajani, Rithesh Kumar, Shubham Jain,
  Jose Sotelo, Aaron Courville, and Yoshua Bengio,
\newblock ``Samplernn: An unconditional end-to-end neural audio generation
  model,''
\newblock {\em arXiv preprint arXiv:1612.07837}, 2016.

\bibitem{model_temporal}
Nicolas Boulanger-Lewandowski, Yoshua Bengio, and Pascal Vincent,
\newblock ``Modeling temporal dependencies in high-dimensional sequences:
  Application to polyphonic music generation and transcription,''
\newblock {\em arXiv preprint arXiv:1206.6392}, 2012.

\bibitem{VAE_paper}
Diederik~P Kingma and Max Welling,
\newblock ``Auto-encoding variational bayes,''
\newblock {\em arXiv preprint arXiv:1312.6114}, 2013.

\bibitem{MuseGAN}
Hao-Wen Dong, Wen-Yi Hsiao, Li-Chia Yang, and Yi-Hsuan Yang,
\newblock ``Musegan: Symbolic-domain music generation and accompaniment with
  multi-track sequential generative adversarial networks,''
\newblock {\em arXiv preprint arXiv:1709.06298}, 2017.

\bibitem{musicvae}
Adam Roberts, Jesse Engel, Colin Raffel, Curtis Hawthorne, and Douglas Eck,
\newblock ``A hierarchical latent vector model for learning long-term structure
  in music,''
\newblock {\em arXiv preprint arXiv:1803.05428}, 2018.

\bibitem{seqgan}
Lantao Yu, Weinan Zhang, Jun Wang, and Yong Yu,
\newblock ``Seqgan: Sequence generative adversarial nets with policy
  gradient.,''
\newblock in {\em AAAI}, 2017, pp. 2852--2858.

\bibitem{VRAE}
Otto Fabius and Joost~R van Amersfoort,
\newblock ``Variational recurrent auto-encoders,''
\newblock {\em arXiv preprint arXiv:1412.6581}, 2014.

\bibitem{GRU}
Kyunghyun Cho, Bart Van~Merri{\"e}nboer, Caglar Gulcehre, Dzmitry Bahdanau,
  Fethi Bougares, Holger Schwenk, and Yoshua Bengio,
\newblock ``Learning phrase representations using rnn encoder-decoder for
  statistical machine translation,''
\newblock {\em arXiv preprint arXiv:1406.1078}, 2014.

\bibitem{empirical_GRU}
Junyoung Chung, Caglar Gulcehre, KyungHyun Cho, and Yoshua Bengio,
\newblock ``Empirical evaluation of gated recurrent neural networks on sequence
  modeling,''
\newblock {\em arXiv preprint arXiv:1412.3555}, 2014.

\bibitem{bachprop}
Florian Colombo and Wulfram Gerstner,
\newblock ``A general model of music composition,''
\newblock {\em arXiv preprint arXiv:1802.05162}, 2018.

\bibitem{DeepBach}
Ga{\"e}tan Hadjeres, Fran{\c{c}}ois Pachet, and Frank Nielsen,
\newblock ``Deepbach: a steerable model for bach chorales generation,''
\newblock {\em arXiv preprint arXiv:1612.01010}, 2016.

\bibitem{Sentence_VAE}
Samuel~R Bowman, Luke Vilnis, Oriol Vinyals, Andrew~M Dai, Rafal Jozefowicz,
  and Samy Bengio,
\newblock ``Generating sentences from a continuous space,''
\newblock {\em arXiv preprint arXiv:1511.06349}, 2015.

\bibitem{boulanger2012modeling}
Nicolas Boulanger-Lewandowski, Yoshua Bengio, and Pascal Vincent,
\newblock ``Modeling temporal dependencies in high-dimensional sequences:
  Application to polyphonic music generation and transcription,''
\newblock {\em arXiv preprint arXiv:1206.6392}, 2012.

\end{thebibliography}

\end{document}